\def\year{2021}\relax
\documentclass[letterpaper]{article} 
\usepackage{aaai21}  
\usepackage{times}  
\usepackage{helvet} 
\usepackage{courier}  
\usepackage[hyphens]{url}  
\usepackage{graphicx} 
\usepackage{natbib}  
\usepackage{caption} 
\usepackage{amssymb}
\usepackage{amsmath}
\usepackage{algorithm}
\usepackage{algorithmic}
\usepackage{tabularx}
\usepackage{mathtools, nccmath}
\usepackage{CJKutf8}
\usepackage[utf8]{inputenc}
\usepackage[T1]{fontenc}
\usepackage[vietnamese,english]{babel}
\usepackage{bbold}
\usepackage{booktabs}
\usepackage{multirow}
\usepackage{siunitx}
\usepackage{microtype}

\newcommand{\bertqa}{{\sc mBERT$_{QA}$}}
\newcommand{\mbert}{{\sc mBERT}}
\newcommand{\eg}{\textit{e.g.}}

\usepackage{todonotes}
\definecolor{deepblue}{rgb}{0.16, 0.32, 0.75}
\definecolor{indiagreen}{rgb}{0.00, 0.44, 0.00}

\title{Multilingual Transfer Learning for QA Using Translation as Data Augmentation}
\author{
    Mihaela Bornea, 
    Lin Pan,
    Sara Rosenthal,
    Radu Florian,
    Avirup Sil
    \\
}
\affiliations{
    IBM Research AI\\
    Thomas J. Watson Research Center, Yorktown Heights, NY 10598\\

    \{mabornea,panl,sjrosenthal,raduf,avi\}@us.ibm.com
   
}

\begin{document}

\maketitle

\begin{abstract}
Prior work on multilingual question answering has mostly focused on using large multilingual pre-trained language models (LM) to perform zero-shot language-wise learning: train a QA model on English and test on other languages. In this work, we explore strategies that improve cross-lingual transfer by bringing the multilingual embeddings closer in the semantic space. 
Our first strategy augments the original English training data with machine translation-generated data. This results in a corpus of multilingual silver-labeled QA pairs that is 14 times larger than the original training set. In addition, we propose two novel strategies, language adversarial training and language arbitration framework, which significantly improve the (zero-resource) cross-lingual transfer performance and result in LM embeddings that are less language-variant. Empirically, we show that the proposed models outperform the previous zero-shot baseline on the recently introduced multilingual MLQA and TyDiQA  datasets.
\end{abstract}

\section{Introduction}
\label{sec:intro} 
Recent advances in open domain question answering (QA) have mostly revolved around machine reading comprehension (MRC)  where the task is to read and comprehend a given text and then answer questions based on it. However, most recent work in MRC has only been in English \eg\ SQuAD \cite{Rajpurkar_2016,rajpurkar2018know}, HotpotQA \cite{yang2018hotpotqa} and Natural Questions \cite{Kwiatkowski2019NaturalQA}. Significant performance gains and the state-of-the-art (SOTA) on these datasets are credited to large pre-trained language models \cite{Devlin2018BERTPO,radford2019language, xlnet}.

\begin{figure}[t]
    \centering
    \small
\framebox{%
  \begin{minipage}{\columnwidth}
    \textbf{Wikipedia Page:} The Reader (2008 film)
\begin{center}
    \textbf{English}
\end{center}
\textbf{Context:} Hanna receives a \textcolor{indiagreen}{life} sentence for her admitted leadership role in the church deaths, while the other defendants are sentenced to four years and three months each.\\ 
\textbf{Question:} What was Hanna's prison sentence?\\
\textbf{Prior work predictions}: \textcolor{red}{four years and three months each}\\
\textbf{This work predictions}: \textcolor{indiagreen}{\textbf{life}}
\begin{center}
    \textbf{Multi-lingual}
\end{center}
\textbf{Context:} 
Hanna es declarada culpable y sentenciada \textcolor{indiagreen}{a cadena perpetua}, mientras que sus compañeras reciben sentencias de cuatro años de cárcel.\\
\begin{CJK*}{UTF8}{gbsn}
\textbf{Question:} Welche Gefängnisstrafe erhielt Hanna?\\
\end{CJK*}
\textbf{Prior work predictions}: \textcolor{red}{cuatro años}\\
\textbf{This work predictions}: \textcolor{indiagreen}{\textbf{a cadena perpetua}}\\
  \end{minipage}}    
  \caption{Examples from the MLQA dataset. Prior zero-shot (ZS) learning models (Lewis \shortcite{lewis2019mlqa}) cannot answer these examples correctly whereas our proposed models (LAF and AT) can (\textcolor{indiagreen}{in green}). }
    \label{fig:mlqa-examples}
\end{figure}

Multilingual BERT (\mbert{}), which is trained on Wikipedia articles from 104 languages and equipped with a 120k shared wordpiece vocabulary, has encouraged a lot of progress on cross-lingual tasks \eg{} XNLI \cite{conneau2018xnli}, NER \cite{keung2019adversarial,wu2019beto} and QA \cite{artetxe2019cross,cui2018span,he2017dureader} by performing \textit{zero-shot} training: train on one language and test on unseen target languages.

In this work, we focus on multilingual QA and, in particular, on two recent large-scale datasets: MLQA~\cite{lewis2019mlqa} and TyDiQA\footnote{All uses of TyDiQA in our paper refer to the Gold Passage task.}~\cite{clark2020tydi}. Both datasets contain English QA pairs but also examples from 13 other diverse languages.

Some examples are shown in Figure \ref{fig:mlqa-examples}. MLQA evaluates two challenging scenarios: 1) \textit{Cross-Lingual Transfer (XLT)}  when the question and the context are in the same language, and 2) \textit{Generalized Cross-lingual Transfer (G-XLT)} when the question is in one language (eg. En) and the context is in another language (eg. De). 
TyDiQA is designed for \textit{XLT} only. Both datasets are challenging for multilingual QA due to the large number of languages and the variety of linguistic phenomena they encompass (e.g. word order, re-duplication, grammatical meanings).

Ideally, we want to build QA systems for all existing languages but it is impractical to collect manually labeled training data for all of them. 
In the absence of labeled data, \cite{clark2020tydi} suggested several research directions for pushing the boundaries in multilingual QA, including zero-shot QA, exploring data augmentation with machine translation, as well as effective transfer learning. These are avenues we explore in our work in addition to asking the following research questions:\\
     
     \noindent \textbf{1. Is a large pre-trained LM sufficient for zero-shot multi-lingual QA?} \\
     Prior work proposes zero-shot transfer learning from English SQuAD data \cite{Rajpurkar_2016} to other languages using \textit{only} a pre-trained LM and competitive results are achieved  on MLQA \cite{lewis2019mlqa} and TyDiQA \cite{clark2020tydi}. We venture beyond zero-shot training by first exploring data augmentation \cite{albert-synth-data} on top of their underlying model. We achieve this by using translation methodologies \cite{yarowsky2001inducing} to augment the English training data. 
     We use machine translation to obtain additional silver labeled data allowing us to improve cross-lingual transfer at a low cost.  Our approach introduces several multilingual extensions to the SQuAD training data: translating just the questions but keeping the context in English, translating just the context but keeping the question in English, and translating the question \textit{and} the context to other languages. This enables us to augment the original English human-labeled training examples with 14 times more multilingual silver-labeled QA pairs.\\
    
    \noindent \textbf{2. Can we bring language-specific embeddings in multi-lingual LMs closer for effective cross-lingual transfer?}\\
    \noindent Our hypothesis is that we can make the cross-lingual QA transfer more effective if we can bring the embeddings in a multilingual pre-trained LM closer to each other in the same semantic space. To answer a question in French it should suffice to train the system on Hindi and not be necessary to train a system on the target language:  hence, French and Hindi should look as if they are the same language.
    We propose two approaches to explore cross-lingual transfer:
        
    In our first approach, we propose a novel strategy based on adversarial training (AT) \cite{miyato2016adversarial,chen2018adversarial,yang2019improving}. We investigate how the addition of a language-adversarial task during QA finetuning for a pretrained LM can significantly improve the cross-lingual transfer performance while causing the embeddings in the LM to become less language-dependent. 
    
    In our second approach, we develop a novel Language Arbitration Framework (LAF) to consolidate the embedding representation across languages using properties of the translation.
        We train additional auxiliary tasks \eg{} making sure an English question and its translation in Arabic produces the same answer when they see the same input context in Spanish. The intuition behind language arbitration is that while we are training the model on English and translated examples, the proposed multi-lingual objectives bring the language-specific embeddings closer to the English embeddings.\\
        
    \noindent Overall, our main contributions in this paper are as follows:\vspace{-0.4em}
        \begin{itemize}\setlength\itemsep{0.0em}
        \item We create a new translation dataset which has \textit{14 times} more multi-lingual 
        silver-labeled QA pairs than SQuAD.
        \item We present an adversarial training approach and a language arbitration framework to bring the LM embeddings closer to each other to improve cross-lingual QA transfer.
        \item We achieve statistically significant improvements compared to prior work \cite{lewis2019mlqa, clark2020tydi} with all of our models.
    \end{itemize}

\section{Multilingual Question Answering}
\label{sec:problem-statement}
\begin{table*}[ht]
\small
\centering
\resizebox{2\columnwidth}{!}{\begin{tabular}{c|cc|cccccccc|c}
\toprule
    \multirow{2}{*}{Dataset} &
      \multicolumn{2}{c|}{Avg. \# of Words} &
          \multicolumn{8}{c|}{Question Type Frequency} &
    \multirow{2}{*}{\# Q-A Pairs} \\
  & Ques. & Ans. &Why &How &What &When &Where &Who &Which &OTHER \\
 \midrule
 SQuAD v1.1 & 10.1 & 3.2 &1,194 &8,082 &37,506 &5,414 &3,261 &8,366 &4,146 &19,630 & 87,599\\
 \midrule
T (Q) & 11.9 & 3.2 &7,164 &48,492 &225,036 &32,484 &19,566 &50,196 &24,876 &117,780 & 525,594\\
T (C) & 10.1 & 4.4 &5,742 &39,668 &190,309 &26,975 &16,225 &42,951 &21,522 &98,298 & 441,690\\
T (Q+C) & 12.1 & 4.4 &5,742 &39,668 &190,309 &26,975 &16,225 &42,951 &21,522 &98,298 & 441,690\\
T(All)  & 11.6 & 4.1 &16,260 &111,664 &530,642 &75,606 &45,494 &119,366 &59,628 &275,116 & 1,233,776\\
\bottomrule
\end{tabular}}
\caption{Comparing our original traning data SQuAD v1.1 with our augmented training data using translation techniques. 
The Question Type is based on the first word in the question.}
\label{tab:translation-data-augmentation}
\end{table*}
In this section, we briefly discuss the LM and QA models. These are the foundations applied to our approach.

\subsection{Pre-trained Language Model}
\label{sec:plm}
Given a token sequence $\mathbf{X} = [x_1, x_2,\ldots,x_T]$, we choose \mbert{}, a deep Transformer \cite{Vaswani_2017} network, which outputs a sequence of contextualized token representations
$\mathbf{H} = [\mathbf{h}_1,\mathbf{h}_2,\ldots,\mathbf{h}_T]$.
\begin{equation}
\mathbf{h}_1,\ldots,\mathbf{h}_T = \mbert{}(x_1,\ldots,x_T)
\label{eq:mbert}
\end{equation}
\noindent \mbert{} has $12$ layers each with $12$ heads and $\mathbf{h}_t\in \mathbb{R}^{768}$. It is pre-trained on $104$ languages and produces SOTA results on many cross-lingual tasks \cite{conneau2018xnli, keung2019adversarial}.

\subsection{Underlying QA model: \bertqa{}}
\label{sec:mbertqa}
We build \bertqa{}, our underlying QA model, as described in \cite{lewis2019mlqa, Devlin2018BERTPO}.
To create the input sequence we concatenate the {\tt[CLS]}, question, {\tt[SEP]} and context tokens.
\bertqa{} adds two dense layers followed by a \emph{softmax} on top of \mbert{} for answer extraction:
\begin{align*}
\boldsymbol{\alpha}_b &= softmax(\mathbf{H} \mathbf{W}_1), \\
\boldsymbol{\alpha}_e &= softmax(\mathbf{H} \mathbf{W}_2),
\end{align*}
\noindent where $\mathbf{W}_1$, $\mathbf{W}_2 \in \mathbb{R}^{768\times 1}$, and $\mathbf{H} \in \mathbb{R}^{T\times 768}$. $\boldsymbol{\alpha}_b^t$ and $\boldsymbol{\alpha}_e^t$ denote the probability of the $t^{th}$ token in the sequence being the answer start and end, respectively. These two layers are trained during the finetuning stage using the cross entropy loss:

\begin{equation}
    \mathcal{L}_{QA} = - \frac{1}{2}(\sum_{t=1}^T \mathbb{1}(\mathbf{b}_t) \log \boldsymbol{\alpha}_{b}^{t} + \sum_{t=1}^T \mathbb{1} (\mathbf{e}_t) \log \boldsymbol{\alpha}_{e}^{t})
\label{eq:qa-loss}
\end{equation}
where $\mathbb{1}(\mathbf{b})$ and $\mathbb{1}(\mathbf{e})$ are one-hot vectors from the ground truth offsets of the answer start and end. \\
\noindent\textbf{Prior work - Zero-shot (ZS) Learning:}
Both \cite{lewis2019mlqa} and \cite{clark2020tydi} propose zero-shot learning for multi-lingual QA by training on English QA data (SQuAD v1.1) and testing on all other languages. This is our basic model and the baseline setting.
We train \bertqa{}
with examples of the form $(Q_{En}, C_{En}, A_{En})$ where $A_{En} \subset C_{En}$. During inference, we use the trained model to extract the answer span $A_y$ from $C_y$ where $y$ is any language.

\section{Models}
\label{sec:models}
In this section, we outline our improvements on top of the prior work on MLQA.

\subsection{Data Augmentation with Translation}
\label{sec:translation}
\begin{algorithm*}[!htb]
\small
\caption{Pseudo-code for adversarial training on the multilingual QA task.}
\begin{algorithmic}[1]
\REQUIRE $\langle Q_{En}^l,C_{En} \rangle, (b,e), L, D, \text{\bertqa{}}, \eta $ .\\
\COMMENT{$Q_{En}^l$ is the translated question (or English), $C_{En}$ English context. $b$ and $e$ are the correct answer start and end positions. $L$ is the language label for the question, $D$ is discriminator, and $h$(\bertqa) is the question representation from the \bertqa{} model, and learning rate $\eta$.}
\FOR{\#epochs}
\FOR{\#steps}
\STATE $\langle Q_{En}^l,C_{En}\rangle_i, (b,e)_i, L_i$ \hspace*{\fill} $\rhd$ Sample in batch of data\\
\STATE $(\boldsymbol{\alpha}_b, \boldsymbol{\alpha}_e)_i \leftarrow$\bertqa$(\langle Q_{En}^l,C_{En}\rangle_i)$ \hspace*{\fill} $\rhd$ QA predictions from \bertqa \\
\STATE $p_{l_i}$ $\leftarrow$ $D$($h$(\bertqa)) \hspace*{\fill} $\rhd$ discriminator language predictions \\
\STATE $\eta \bigtriangledown_{\theta_{QA}} (\mathcal{L}_{QA} + \mathcal{L}_{adv})$ \hspace*{\fill} $\rhd$ update QA model with Eq \ref{eq:qa-loss} and \ref{eq:adv-loss}\\
\STATE $p_{l_i}$ $\leftarrow$ $D$($h$(\bertqa)) \hspace*{\fill} $\rhd$ discriminator language predictions \\
\STATE $\eta \bigtriangledown_{\theta_{D}} \mathcal{L}_{D}$ \hspace*{\fill} $\rhd$ update Discriminator with Eq \ref{eq:discriminator-loss}

\ENDFOR
\ENDFOR
\end{algorithmic}
\label{gan-algorithm}
\end{algorithm*}

\begin{algorithm*}[t]
\small
\caption{Pseudo-code for our language arbitration framework for the multilingual QA task.}
\begin{algorithmic}[1]
\REQUIRE  $\langle Q_{En},C_{En}\rangle , \langle Q_{En}^l,C_{En}\rangle , (b,e), \text{\bertqa{}},\eta$
\COMMENT{$Q_{En}^l$ is the translated question, $C_{En}$ English context. $b$ and $e$ are correct answer start and end, $h$(\bertqa) is the question representation from the \bertqa{} and $\eta$ is the learning rate .}
\FOR{\#epochs}
\FOR{\#steps}
\STATE $\langle Q_{EN},C_{En}\rangle_i$, $\langle Q_{En}^l,C_{En}\rangle_i, (b,e)_i$ \hspace*{\fill} $\rhd$ Sample in batch of data\\
\STATE $(\boldsymbol{\alpha}_b^{En}, \boldsymbol{\alpha}_e^{En})_i\leftarrow $\bertqa$(\langle Q_{En},C_{En}\rangle_i )$ \hspace*{\fill} $\rhd$ generate predictions from the QA model for En\\
\STATE$(\boldsymbol{\alpha}_b^l, \boldsymbol{\alpha}_e^l)_i\leftarrow$\bertqa$(\langle Q_{En}^l,C_{En}\rangle_i )$ \hspace*{\fill} $\rhd$ generate predictions from the QA model on language $l$\\
\STATE $\eta \bigtriangledown_{\theta_{QA}} (\mathcal{L}_{QA}^{En}+\mathcal{L}_{QA}^{l})$ \hspace*{\fill} $\rhd$ update QA model with Eq \ref{eq:qa-loss} for En and language $l$\\
\STATE $(\boldsymbol{b^{En}}, \boldsymbol{e^{En}})_i\leftarrow argmax((\boldsymbol{\alpha}_b^{En}, \boldsymbol{\alpha}_e^{En})_i)$ \hspace*{\fill} $\rhd$ find the begin and end of answer when question in En\\
\STATE $\eta \bigtriangledown_{\theta_{QA}} \mathcal{L}_{PSA}$ \hspace*{\fill} $\rhd$ update QA model with with the PSA loss in Eq \ref{eq:psa-loss} \\
\STATE $(\overline{h}_{Q_{En}})_i\leftarrow h($\bertqa{}$(\langle Q_{En},C_{En}\rangle_i ))$ \hspace*{\fill} $\rhd$ get question representation for En\\
\STATE $(\overline{h}_{Q_{En}^l})_i \leftarrow h($\bertqa{}$(\langle Q_{En}^l,C_{En}\rangle_i ))$ \hspace*{\fill} $\rhd$ get question representation for language $l$\\
\STATE $\eta \bigtriangledown_{\theta_{QA}} \mathcal{L}_{QS}$ \hspace*{\fill} $\rhd$ update QA model with the QS loss in Eq \ref{eq:QS-loss} \\
\ENDFOR
\ENDFOR
\end{algorithmic}
\label{mtl-algorithm}
\end{algorithm*}

\noindent Our first approach beyond zero-shot QA is to introduce data-augmentation \cite{yu2018qanet, albert-synth-data} based models. 
Since we only have English examples to train our system on, we expand our training data and explore several translation-based data augmentation models for MLQA. Table~\ref{tab:translation-data-augmentation} shows statistics for the different datasets. 
We use the IBM Watson Language Translator  \cite{translation_api} to:\\
\noindent\textbf{1. Translate (Q+C)}: 

We pick a language $l \in L$ where $L =\{De, Es, Ar, Hi, Zh\}$\footnote{Our translation api does not support Vietnamese and Swahili.} and translate $(Q_{En}, C_{En}, A_{En})$ to create examples $(Q_{En}^l, C_{En}^l, A_{En}^l)$ in that language. We do this for each of the 5 languages. Note, $Q_{En}^l$ and $C_{En}^l$ are the translations of $Q_{En}$ and $C_{En}$ and $A_{En}^l \in C_{En}^l$ is the translated answer, all in language $l$.
In order to obtain the alignment of the gold answer $A_{En}$ in the translated context $C_{En}^l$, we place pseudo HTML tags around $A_{En}$ and then translate $C_{En}$.
Note that the main challenge of this strategy is the answer alignment step\footnote{Dev experiments suggests that this is better than using the translation alignment scores.} and we only keep the translated examples where this succeeds. 
The number of translated examples we obtained is $87,062$ for German, $77,759$ for Spanish, $84,185$ for Arabic, $20,981$ for Hindi and $84,104$ for Chinese. The final data set including English has $441,690$ examples. 
The percentage of reduced question type ranges from $14\%$ (Which) to $20\%$ (Why).  \\
\noindent\textbf{2. Translate(Q)} : \textit{Only} $Q_{En}$ is translated to other languages leaving $C_{En}$ intact to create examples $\cup_l(Q_{En}^l, C_{En}, A_{En})$.
This data augmentation strategy produces a more accurate dataset since it does not require the answer alignment stage which can be error-prone. We translate every $Q_{En}$ to 5 other languages and we obtain a dataset of $525,594$ examples, which is 6 times larger than SQuAD v1.1. T(Q) increases the average number of words in the question by $1.8$.\\
\noindent\textbf{3. Translate (C)}: We \textit{only} translate $C_{En}$ to other languages to create $\cup_l(Q_{En}, C_{En}^l, A_{En}^l)$. We use the same answer alignment strategy as in \textit{Translate (Q+C)} to generate the gold answer $A_{En}^l$ for the translated examples in $L$. We obtain $441,690$ examples (same as Translate (Q+C)). T(C) increases the average number of words in the answer by 1.2.\\
\noindent\textbf{4. Translate(ALL)}: We combine the data from all the 3 strategies together to create a meta-translation model with $1,233,776$ examples, 14 times larger than SQuAD.

\subsection{Adversarial Training}
\label{sec:adversarial}
Translation-based strategies provide ample scope for \bertqa{} to train on plenty of $\{Q,C,A\}$ examples where $Q$ and $C$ can be in different languages. However, it can still be challenging as new languages can continuously be added to the model requiring optimal MT systems in all languages. Therefore, it is important to explore bringing the embeddings of different languages in \mbert{} close to each other to achieve effective cross-lingual transfer. For this purpose, we introduce a novel multilingual adversarial training (AT) method inspired by \cite{NIPS2014_5423}.
The goal is to fine-tune \mbert{} so that its embeddings become as \emph{language-invariant} as possible. Algorithm \ref{gan-algorithm} provides an overview of this approach.
 
Concretely, we use the \textbf{Translate(Q)} strategy outlined in the previous section, and for every $\{Q_{En}, C_{En}, A_{En}\}$, we derive examples of $\{Q_{En}^l, C_{En}, A_{En}\}$, where the question is translated. All the examples are added to the training data. The discriminator $D$ of the AT model is trained to classify the question representation in different languages to the correct language label $L \in [En, De, Es, Ar, Hi, Zh]$. 
We use the {\tt[CLS]} token to derive a single question representation as input for $D$ and train with cross-entropy loss:
\vspace{-0.3em}
\begin{equation}
\mathcal{L}_{D} = -\sum_{l=1}^L \mathbb{1}(\mathbf{g}_{l}) \log \mathbf{p}_{l},
\label{eq:discriminator-loss}
\end{equation}
where $\mathbb{1}(\mathbf{g}$) is a one-hot vector for the ground-truth language labels, and $\mathbf{p}$ are the language predictions from the model. $D$ is implemented as a multilayer perceptron.

Under the AT objective, the underlying QA model, in addition to the QA objective, is trained to also minimize the KL-divergence between the uniform distribution, and the language labels predicted by the discriminator.
\vspace{-0.4em}
\begin{equation}
\mathcal{L}_{adv} = -\sum_{l=1}^L  KL[U(\mathbf{g}_l) || \log \mathbf{p}_{l}],
\label{eq:adv-loss}
\end{equation}
\noindent $\mathcal{L}_{adv}$ encourages the LM embeddings to appear uniform to the discriminator, across all languages. In contrast, $\mathcal{L}_{D}$ drives the discriminator to recognize the language.
During training, in each step, we first update \bertqa{} with $\mathcal{L}_{QA} + \mathcal{L}_{adv}$ (See Eq \ref{eq:qa-loss} for $\mathcal{L}_{QA}$) while fixing the parameters of the discriminator (Alg. \ref{gan-algorithm} line 6), and then update the discriminator with $\mathcal{L}_{D}$ fixing those of \bertqa{} (line 8).

In addition to performing AT using all $6$ languages, \textbf{AT (en-all)}, we also conduct experiments picking just one random language (\eg{} $l=Zh$) to perform \textbf{AT (en-zh)}. 
\subsection{Language Arbitration Framework}
\label{sec:mtl} 
In this section, we explore an alternative approach for bringing the language-specific embeddings closer to each other using a novel Language Arbitration Framework (LAF) to train a multilingual QA model. Just like a regular human \textit{arbitrator}, LAF's job at the end of training is to make sure the same question in different languages produce the same answer while maintaining that the underlying representation of the questions are the same.
Similar to the AT method,  \textbf{Translate(Q)} is used to generate our training examples.
For every $\{Q_{En}, C_{En}, A_{En}\}$ in the original English dataset, we derive an augmented training set with example pairs  $(\{Q_{En}, C_{En}, A_{En}\},\{Q_{En}^l, C_{En}, A_{En}\})$ where the question is translated to language $l \in L$. Training of LAF proceeds with such example pairs and exploits properties of the translation to consolidate the LM embeddings.  In addition to training the base \bertqa{} model on English and the translation, using the standard objective from Eq~(\ref{eq:qa-loss}), LAF also performs the following objectives during training:\\
\textbf{1. Produce the same answer (PSA):} PSA encourages the translation $\langle Q_{En}^{l}, C_{En} \rangle$ to produce the same answer as the original example $\langle Q_{En}, C_{En} \rangle$, for all languages $l \in L$.  We run \bertqa{} on English and the translation. Then, in addition to computing $\mathcal{{L}}_{QA}$ (Equation \ref{eq:qa-loss}) we compute the additional loss:

    \begin{equation}
    \mathcal{L}_{PSA} = - \frac{1}{2}(\sum_{t=1}^T \mathbb{1}(\mathbf{b}_{t}^{En}) \log \boldsymbol{\alpha}_b^{l^t} + \sum_{t=1}^T \mathbb{1} (\mathbf{e}_{t}^{En}) \log
    \boldsymbol{\alpha}_e^{l^t})
    \label{eq:psa-loss}
    \end{equation}    
    
\noindent$\mathbb{1}(\mathbf{b}_{t}^{En})$ and $\mathbb{1}(\mathbf{e}_{t}^{En})$  are one-hot vectors indicating the answer start and end positions predicted by the \bertqa{} for $\langle Q_{En}, C_{En} \rangle$. 
$\boldsymbol{\alpha}_b^{l}$ and $\boldsymbol{\alpha}_e^{l}$ denote the answer begin and end probability predicted by \bertqa{} for $\langle Q_{En}^{l}, C_{En} \rangle$. While \textbf{Translate(Q)} optimizes the standard $\mathcal{L}_{QA}$ objective on translated data, $\mathcal{L}_{PSA}$ uses the English predictions for additional supervision and brings the LM embeddings closer by maintaining agreement between English and the translation. This is beneficial in cases where there is partial overlap between the English predicted answer and the gold label.

\noindent \textbf{2. Produce the same answer and question similarity (PSA+QS):} In this approach, in addition to the PSA loss, we also compute the \textit{cosine-similarity} between $Q_{En}$ and $Q_{En}^l$ in all languages $l \in L$. The intuition is that the cosine similarity of translations should be high, encouraging the embeddings to move even closer to each other.
    
To obtain a single question representation, $\overline{h}_{Q_{En}}$ for $En$ and $\overline{h}_{Q_{En}^l}$ for language $l$, we perform average pooling over the hidden vectors for the question tokens from \mbert{}.
    \begin{equation}
    \mathcal{L}_{QS} = 1 - cosine(\overline{h}_{Q_{En}},\overline{h}_{Q_{En}^l})
\label{eq:QS-loss}
\end{equation}

\noindent In addition to performing PSA and PSA+QS in \textit{all} the languages, we also apply them in a single language, $l=Zh$ as \textbf{PSA(en-zh)} and \textbf{PSA+QS(en-zh)}.

\section{Experiments}
\label{sec:experiments}

\subsection{Data and Evaluation Metric}
\textbf{MLQA:} We first evaluate our techniques on MLQA \cite{lewis2019mlqa} which is a large multilingual QA dataset that covers 7 languages as listed in Table~\ref{tab:results}. The dataset is 4-ways language-parallel with parallel passages from Wikipedia articles on the same topic. Questions are originally asked in English and they are translated to other target languages. 

The dataset provides a development set (1,148 parallel instances) that is significantly smaller than the blind test (11,590 parallel instances). Hence, we train our models on the SQuAD v1.1 dataset (details in Table \ref{tab:translation-data-augmentation}). We also create a much larger multi-lingual training corpus, as outlined in Sec \ref{sec:translation}, with the help of machine translation. To provide a comprehensive evaluation of our techniques we run all experiments on the MLQA dataset since it was designed for both G-XLT and XLT task.

\noindent\textbf{TyDiQA:} We choose the best models based on our MLQA experiments and run them on the TyDiQA
\cite{clark2020tydi} GoldP dataset. The GoldP task was designed only for XLT evaluation and is similar to MLQA.  
There are 9 languages of which English (en) and Arabic (ar) are the only ones in common between TyDiQA and MLQA. 
Although TyDiQA has a multilingual training set, in this work we train our models on SQuAD v1.1 in order to test the cross-lingual transfer ability of our proposed models. We also create a separate training corpus by translating the questions to the TyDiQA languages, resulting in 700,792 examples. We use this augmented training corpus to implement AT and LAF. The evaluation (dev) set contains 5,077 instances.

\noindent\textbf{Evaluation Metric:} We use the official evaluation metric from both datasets and report the mean token F1 \footnote{We report token-level F1 as opposed to Exact Match (EM) as the latter severely penalizes a system if it adds functions words.}.
For MLQA, we report separate F1 scores on both the G-XLT and XLT tasks. For TyDiQA, we report the XLT F1 since the question is always in the language of the context.

\subsection{Hyper-parameters}
We perform hyper-parameter selection on the SQuAD and MLQA dev split. We use $3\times 10^{-5}$ as the learning rate, $384$ as maximum sequence length, and a doc stride of $128$. 
Everything except ZS was trained for 1 epoch. We use the same hyper-parameter values on the MLQA test set and TyDiQA experiments. The best question representation is achieved with the {\tt[CLS]} token for AT and average pooling for LAF (PSA+QS). Other methods tried were the concatenation of {\tt[CLS]} and {\tt[SEP]}.
The discriminator is implemented as a multilayer perceptron with $2$ hidden layers and a hidden size of $768*4$.   
For both AT and LAF, in addition to (en-zh), which was chosen at random, we also experimented with German, the language closest to English. Both achieve similar performance.

\begin{table*}[t]
\tiny
\centering
\resizebox{2\columnwidth}{!}{\begin{tabular}{c|c|ccccccc|c|c}
\toprule
      \multirow{2}{*}{Model} &
      \multirow{2}{*}{Method} &
      \multicolumn{7}{c|}{MLQA Languages (G-XLT) } &
    \multirow{2}{*}{G-XLT} &
    \multirow{2}{*}{XLT} \\
  &&ar 	&de 	&en 	&es 	&hi 	&vi 	&zh & &\\
 \midrule
\bertqa{} & ZS & 46.9 & 51.4 & 60.2 & 55.0 & 47.0 & 52.0 & 49.7 & 51.7 ($\pm$0.4) & 61.7 ($\pm$0.3)
\\\cmidrule{1-11}
\multirow{4}{*}{Trans} & T(Q) & \textbf{53.8} & 60.8 & 73.5 & 65.4 & \textbf{53.2} & 63.2 & 56.7 & 60.9 ($\pm$0.2) & \textbf{64.9} ($\pm$0.2)\\
& T(C) & 44.8 & 51.7 & 62.0 & 57.6 & 42.7 & 55.8 & 50.8 & 52.2 ($\pm$1.0) & 58.5 ($\pm$0.9)\\
 & T(Q+C) & 48.9 & 58.4 & 70.4 & 63.6 & 46.8 & 61.4 & 54.4 & 57.7 ($\pm$0.1) & 64.3 ($\pm$0.0)\\
 & T(ALL) & 52.6 &\textbf{61.1} &\textbf{73.8} &\textbf{66.3} &50.6 &\textbf{64.8} &\textbf{58.2} &\textbf{61.1} ($\pm$ 0.1) &64.2 ($\pm$0.2)
\\\cmidrule{1-11}
 \multirow{2}{*}{AT} & (en-zh) & 50.5 & 56.7 & 68.0 & 60.8 & 50.7 & 57.4 & 51.5 & 56.5 ($\pm$ 0.1) & 62.8 ($\pm$ 0.1)\\
 & (en-all) & \textbf{54.1} & \textbf{61.1} & \textbf{73.6} & \textbf{65.5} & \textbf{54.2} & \textbf{63.4} & \textbf{56.8} & \textbf{61.2} ($\pm$ 0.1) & \textbf{65.2} ($\pm$ 0.1)
\\\cmidrule{1-11}
 \multirow{4}{*}{LAF} & PSA (en-zh) & 50.8 & 56.9 & 68.6 & 61.3 & 51.0 & 57.8 & 51.6 & 56.9 ($\pm$ 0.1) & 62.8 ($\pm$0.1)\\
 & PSA+QS (en-zh) & 50.7 & 56.6 & 68.5 & 61.2 & 51.0 & 57.7 & 51.8 & 56.8 ($\pm$ 0.4) & 62.7 ($\pm$0.1)\\
 & PSA (en-all) & 54.5 & 61.4 & 74.2 & 66.0 & 54.6 & 64.2 & 57.5 & 61.8 ($\pm$0.1) & 65.6 ($\pm$0.0)\\
 & PSA+QS (en-all) & \textbf{54.8} & \textbf{61.5} & \textbf{74.3} & \textbf{66.1} & \textbf{54.9} & \textbf{64.3} & \textbf{57.6} & \textbf{61.9} ($\pm$0.1) & \textbf{65.7} ($\pm$0.0)\\
\bottomrule
\end{tabular}}
\caption{Our results on MLQA test averaged over 3 runs. We compare our models against the previous baseline \cite{lewis2019mlqa}: ZS setting with \bertqa{}. Best numbers within the method are in bold. The best LAF and AT models are statistically significantly better than the best Trans model.}
\label{tab:results}
\end{table*}

\subsection{MLQA Results}
Table \ref{tab:results} shows the performance of various competing strategies for MLQA. 
For each language of the context we report the G-XLT performance averaged across questions in all the 7 languages. The final two columns show the \textit{overall} G-XLT and the XLT performance across all the 7 languages.\\
\textbf{Zero-shot:} We report the results of our re-implementation of the ZS setting of \bertqa{} \cite{lewis2019mlqa} which is the underlying QA model and show our improvements on top it.\\

\textbf{Translation:} T(Q) provides the biggest improvement out of all the competing translation techniques T(C), T(Q+C) with an overall gain (on average) of 6 points on G-XLT and 3.5 points on XLT. We believe that this degradation is due to answer alignment errors when translating the context. The alignment also causes a loss in training examples compared to the case when just the questions are translated.
Note that the T(C) model is the weakest as it is the most affected by the alignment strategies and has the highest standard deviation among all the models. Combining all the strategies together provides a tiny improvement on G-XLT but at a cost to XLT performance: we believe that the T(C) data hurts this model and the parameters of \mbert{} alone are not sufficient to bring embeddings of different languages close to each other even with translation data. As we add more languages, the per-language capacity of the QA system decreases. This impacts the performance (known as the \textit{curse of multilinguality}~\cite{conneau2019unsupervised}).\\
\textbf{Adversarial Training:} 
We first experiment with the AT (en-zh) model and noticed that adding a single language to the training data significantly improves performance over ZS. However AT (en-zh) is not strong \{56.5 (G-XLT), 62.8 (XLT)\} compared to T(Q), T(Q+C) and T(All). During training the discriminator is tasked to make a binary classification between En and Zh in this case. We hypothesize that this task may be too easy to balance the overall system training, since \cite{sonderby2016amortised} showed that making the discriminator work harder is beneficial for training AT models.
We leave training AT individually with each of the 6 other languages as part of our future work.
When we extend the scope of the model to look at all languages together, we get the best performing MLQA system so far with \{61.2 (G-XLT), 65.2 (XLT)\}.\\
\textbf{Language Arbitration Framework:} Similar to AT, for LAF, we first start with an `en-zh' model and then move on to an `en-all' model. Our PSA+QS is weaker than just doing PSA on `en-zh' suggesting again that choosing only one extra language in the LAF setting improves over the ZS baseline but is not as beneficial as adding all languages together.
By choosing all the languages, we get the best performing overall model on the test split. PSA (en-all) does not lag behind but PSA+QS (en-all) provides an overall improvement of 10.2 and 4 points and 0.8 and 1.5 points improvement in G-XLT and XLT respectively over the ZS baseline and the best translation system `T(All)'. It is more beneficial to bring the multilingual embeddings closer to English for LAF than the global level as in the AT approach. \\ 
We observe that the best LAF model is consistently better than the competing strategies for \textit{all} language pairs: 61.9 vs 61.1 (G-XLT) and 65.7 vs. 64.2 (XLT). Table \ref{tab:mtl_psa+qa} shows the detailed results of our best LAF model across all MLQA language combinations. In Table \ref{tab:zs-vs-mtl}, we compare our best performance on XLT against ZS results introduced in prior work \cite{lewis2019mlqa} achieving a significant 4 point improvement\footnote{Note that our ZS re-implementation results in higher numbers than Table 5 in \cite{lewis2019mlqa}.}.

\noindent\textbf{Statistical Significance:}
We compute statistical significance via the Fisher randomization test. The best LAF model (PSA+QS(en-all)) is 
statistically significantly better than the best AT and Translation model ($p<0.05$). The best model for all three methods (T(Q), AT (en-all) and PSA+QS (en-all)) 
is significantly better than the ZS baseline.

\begin{table}[t]
\small
\center
\resizebox{\columnwidth}{!}{\begin{tabular}{l|lllllll|l}
\toprule
q$\backslash$c & ar   & de   & en   & es   & hi   & vi   & zh  & \textbf{AVG}\\
\midrule
ar   & 58.0 & 59.7 & 70.3 & 62.7 & 51.1 & 61.6 & 54.0 & 59.6 \\ 
de   & 58.6 & 65.5 & 78.0 & 71.0 & 59.0 & 66.2 & 59.9 & 65.5 \\ 
en   & 56.8 & 64.9 & 80.2 & 69.6 & 56.6 & 66.6 & 59.6 & 64.9 \\ 
es   & 55.8 & 66.0 & 77.7 & 70.2 & 55.3 & 64.5 & 58.2 & 64.0 \\ 
hi   & 50.5 & 57.1 & 70.3 & 61.5 & 58.9 & 60.7 & 54.0 & 59.0 \\ 
vi   & 49.3 & 56.7 & 69.0 & 61.4 & 50.8 & 64.0 & 54.7 & 58.0 \\ 
zh   & 54.3 & 60.9 & 74.3 & 66.0 & 52.4 & 66.3 & 63.1 & 62.5 \\
\midrule
\textbf{AVG} & 54.8 & 61.5 & 74.3 & 66.1 & 54.9 & 64.3 & 57.6 & 61.9\\
\bottomrule
\end{tabular}}
\caption{\textbf{G-XLT} F1 scores of the LAF:PSA+QS (en-all) model on the overall test set for individual cross languages performance. \textbf{XLT} F1 is 65.7 averaged across the diagonal, as shown with the G-XLT results in the last row of Table \ref{tab:results}.}
\label{tab:mtl_psa+qa}
\end{table}

\begin{table}
\small
\centering
    \resizebox{\columnwidth}{!}{\begin{tabular}{lcccccccc}
    \toprule
         \textbf{Model} &\textbf{ar} &\textbf{de} &\textbf{en} &\textbf{es}
         &\textbf{hi} &\textbf{vi} &\textbf{zh}  & \textbf{XLT}\\
         \midrule
         ZS &51.7	&60.6	&80.4	&66.8	&50.5	&61.4	&60.1	&61.7\\
        LAF	&\textbf{58.0}	&\textbf{65.5}	&\textbf{80.2}	&\textbf{70.2}	&\textbf{58.9}	&\textbf{64.0}	&\textbf{63.1} &\textbf{65.7}\\
         \bottomrule
    \end{tabular}}
    \caption{XLT F1 scores of ZS and LAF with \mbert{}.}
    \label{tab:zs-vs-mtl}
\end{table}

\begin{table}[h]
\centering
\setlength{\tabcolsep}{3pt}
\resizebox{\columnwidth}{!}{\begin{tabular}{c|ccccccccc|c}
\toprule
      \multirow{2}{*}{Model} &
      \multicolumn{9}{c|}{TyDiQA Languages} &
      \multirow{2}{*}{XLT} \\
  &en 	&bn 	&ko 	&in 	&te 	&sw 	&ar & ru & fi & \\
 \midrule
ZS & \textbf{75.0} & 62.8 & 55.3 & 61.6 & 49.9 & 57.8 & 61.6 & 65.0  & 58.5  & 60.8($\pm1.0$)
\\ \midrule
T(Q)* &  73.2 & 59.4 & 56.7 & 61.4 & 47.8 & 62.4 & 67.5 & 63.8 & 54.6 & 60.8($\pm0.3$) 
\\
AT*   & 74.1 & 59.9 & 56.5 & 63.0 & 49.0 & 63.8 & 67.0 & 64.6 & 56.7 & 61.6($\pm0.9$)
\\
 LAF*  & 74.1 & 59.9 & 55.3 & 64.1 & 49.1 & 63.8 & \textbf{68.4} & 65.9 &  57.3 & 61.9($\pm0.4$)
\\
\midrule
T(Q) &  73.7 & 63.8 & 59.7 & 70.8 & 49.5 & 60.6 & 65.5 & 65.7 & 69.3 & 64.3($\pm0.3$) 
\\
AT & 73.7 & 64.2 & \textbf{62.1} & 71.9 & 49.1 & 62.2 & 66.6 & 66.2 & 70.8 & 65.2($\pm0.5$)
\\
LAF   & 74.3 & \textbf{67.6} & 61.9 & \textbf{72.0} & \textbf{50.6} & \textbf{62.4} & 68.0 & \textbf{67.0} & \textbf{71.2} & \textbf{66.1}($\pm0.5$)\\
\bottomrule
\end{tabular}}
\caption{Our results on TydiQA dev. We compare our models against the previous baseline \cite{clark2020tydi}: ZS setting with \bertqa{}. T(Q)*, AT*, LAF* are the MLQA models. The LAF* and AT* are statistically significantly better than ZS. The LAF and AT models are statistically significantly better than the T(Q) model and ZS.}
\label{tab:results_tydi}
\end{table}

\subsection{TyDiQA Results}
Table~\ref{tab:results_tydi} shows the results on TyDiQA. We first experiment with the same models that we trained for MLQA by translating SQuAD to the MLQA languages. In this setting, we evaluate cross-lingual transfer beyond translation, since \textit{en} and \textit{ar} are the only languages the two datasets have in common. Our best MLQA translation strategy T(Q), improves the F1 significantly on \textit{ar} but it is slightly detrimental for the other target languages. On average the translation baseline shows no improvement over ZS. The best performing model is LAF with 1.5 F1 gains over ZS. LAF also has the best cross-lingual transfer performance, improving Indonesian (in), Swahili (sw), Russian (ru) as well as \textit{ar} compared to the ZS baseline.
We also tested our models trained by translating SQuAD to the TyDiQA languages. 
In this case, we notice consistent trends with the MLQA results. All techniques improve the cross-lingual transfer across all languages.  Data augmentation with MT shows large improvement over ZS increasing the F1  by 3.4 points. AT is better compared to T(Q) and the best results are obtained with cross-lingual LAF with an average increase of 5.3 F1 points compared to ZS. Our improvements over ZS and T(Q) are statistically significant and we used the Fisher randomization test.

\subsection{Error Analysis}
We take a random sample of our dev data and perform error analysis on the output to provide insights into our contributions. The correct answer predicted by the better model is shown in \textcolor{indiagreen}{\underline{\textbf{green}}} and the incorrect answer predicted by the poorer model is shown in \textcolor{red}{\textit{red}}.\\
\noindent\textbf{Translation is better than ZS:}\\
\textbf{C(En):} \textcolor{red}{\textit{Stephen William Kuffler}} is known for his research on neuromuscular junctions in frogs, presynaptic inhibition, and the neurotransmitter \textcolor{indiagreen}{\underline{\textbf{GABA}}}.\\
\begin{CJK*}{UTF8}{gbsn}
\textbf{Q(Zh):} 他以什么神经递质的名字而闻名\\
\end{CJK*}
\textbf{Explanation:} Data augmentation helps.\\

\noindent\textbf{AT is better than Translation:}\\
\textbf{C(De):} Heftiger Regen verursachte auf \textcolor{indiagreen}{\underline{\textbf{Hawai'i}}} geringere Sch\"{a}den durch \"{o}rtliche \"{U}berflutungen ..\textcolor{red}{\textit{auf der Nordhalbkugel}} die st\"{a}rksten Winde und...\\
\textbf{Q(En):} Where were heavy rains?\\
\textbf{Explanation:} Adversarial training makes the \mbert{} embeddings more language-invariant.\\

    \noindent\textbf{LAF is better than AT:}\\
\textbf{C(Es):} La película, que combina animación por computadora con acción en vivo, fue dirigida por \textcolor{red}{\textit{Michael Bay}}, con \textcolor{indiagreen}{\underline{\textbf{Steven Spielberg}}} como productor ejecutivo.\\
\textbf{Q(Vi):} \foreignlanguage{vietnamese}{Ai là đạo diễn sản xuất bộ phim Transformers năm 2007?}\\
\textbf{Explanation:} LAF makes the \mbert{} embeddings even more language-invariant than AT.\\\\
\textbf{LAF $\&$ AT are better than Translation:}\\
\textbf{C(En):} Berlin is a world city of \textcolor{indiagreen}{\underline{\textbf{culture, politics, media and science}}}...serves as a continental hub...metropolis is a popular \textcolor{red}{\textit{tourist destination}}.\\
\textbf{Q(De)}: Wofür war Berlin bekannt?\\
\textbf{Explanation:} See previous explanations.

\section{Related Work}
\label{sec:related-work}
A large number of recent QA/ MRC datasets such as SQuAD \cite{Rajpurkar_2016,rajpurkar2018know}, TriviaQA \cite{DBLP:journals/corr/JoshiCWZ17}, NewsQA \cite{newsqa} and Natural Questions \cite{Kwiatkowski2019NaturalQA} have focused on English and have not explored multilingual QA.

There are plenty of non-English QA datasets \cite{gao2016multilingual,he2017dureader,shao2018drcd, mozannar2019neural,gupta2018mmqa,lee2018semi, li2018multi, asai2018multilingual, croce2019enabling} in Chinese, Arabic, Hindi, Korean, French, Japanese and Italian. These datasets are 2-3 way parallel or mono-lingual. XQuAD \cite{artetxe2019cross} is a translated subset of SQuAD v1.1 into 10 languages.  
The most competitive multi-lingual datasets are MLQA and TyDiQA due to their scale and use of the original contexts as they appear in Wikipedia rather than manual translation from English.

Prior work has explored (back)-translation for data-augmentation \cite{yu2018qanet}, multi-task learning \cite{mccann2018natural,bonadiman2017multitask,chen2017reading}, adversarial learning \cite{wallace2019universal,yang2019improving,wang2018robust,zhu2019freelb, keung2019adversarial, chen2018adversarial} either for mono-lingual QA or for other NLP tasks. None of these have explored multi-lingual techniques similar to ours that make the embeddings in the LM become language-agnostic.

Contrary to our approach, \cite{yuan2020enhancing} present results on MLQA but assume access to a commercial search engine as well as web queries to create their specialized training data for their answer boundary detection task. They only report XLT results on 3/7 MLQA languages, whereas, we evaluate on \textit{all} 7 languages and report \textit{both} XLT and G-XLT performance. We also note that access to a search engine is not always feasible and since the authors do not provide the web queries it is unclear how to extend their technique to other languages. 

Perhaps, the closest work to ours is \cite{cui2019cross}, their approach relies on back-translation and an ensemble of two QA systems one on source (context) and one on target (question) language. Our proposed methods 1. do not rely on back-translation, 2. we introduce more diverse translation models and 3. we introduce two novel strategies for multi-lingual QA based on language arbitration and adversarial learning. Most importantly their ensemble approach relies on training data in the target language whereas we do not.

Choosing which of the multilingual LMs (e.g. \mbert{} \cite{Devlin2018BERTPO}, XLM-R \cite{conneau2019unsupervised} and M4 \cite{arivazhagan2019massively})  to use for MLQA is a separate thread of work that involves comparing pre-training objectives and which large corpora to train on and is not the main focus of this paper. Due to the large number of experiments we ran we focus on one framework and we chose \mbert{}.

\section{Conclusion}
\label{sec:conclusion}
In this work, we highlight open challenges in the existing multilingual approach by \cite{lewis2019mlqa} and \cite{clark2020tydi}. Specifically, we show that large pre-trained multi-lingual LMs are not enough for this task. We produce  several novel strategies for multilingual QA that go beyond zero-shot training and outshine the previous baseline built on top of \mbert{}. We present a translation model that has \textit{14 times more} training data. Further, our AT and LAF strategies utilize translation as data augmentation to bring the language-specific embeddings of the LM closer to each other. These approaches help us significantly improve the cross-lingual transfer. Empirically, our models demonstrate strong results and all approaches improve over the previous ZS strategy. We hope these techniques
spur further research in the field such as exploring other multilingual LMs and invoking additional networks on top of large LMs for multilingual NLP. 

\section*{ Acknowledgments}
We thank Graeme Blackwood for his help with the machine translation api.
We are grateful to Salim Roukos and the IBM MNLP team for the helpful discussions.
We also thank the anonymous reviewers for their suggestions that helped us improve this paper.

\bibliography{ms}

\end{document}